%% file: main.tex
\documentclass[11pt]{article}
\usepackage{geometry}
\usepackage{coling2020}
\usepackage{times}
\usepackage{url}
\usepackage{latexsym}
\usepackage{microtype}

\colingfinalcopy 

\usepackage{tabularx}
\usepackage{import}
\usepackage{graphicx}
\usepackage{amsmath}
\usepackage{bbm}
\usepackage{makecell}
\usepackage{array}
\newcolumntype{?}{!{\vrule width 1.5pt}}

\usepackage{adjustbox}
\usepackage{import}
\usepackage{color} 
\usepackage[multiple]{footmisc}
\usepackage[normalem]{ulem} 
\usepackage{enumitem}
\usepackage{booktabs} 
\usepackage[verbose]{wrapfig} 
\usepackage{pifont} 
\newcommand{\descrcell}[2]{%
  \scriptsize
  \begin{tabular}[t]{@{}c@{}}\normalsize#1\\#2\end{tabular}%
}
\newcolumntype{H}{>{\setbox0=\hbox\bgroup}c<{\egroup}@{}} 
\usepackage{subcaption}
\usepackage[title]{appendix}

\definecolor{darkGreen}{rgb}{0,0.6,0}

\newcommand{\nh}[1]{\textcolor{blue}{[NH - #1]}}

\newcommand{\usage}[1]{{\color{red} {#1}}} 

\newcommand{\hide}[1]{}

\title{SemEval-2020 Task 7:  Assessing Humor in Edited News Headlines}

\author{Nabil Hossain$^\dagger$, John Krumm$^\ddagger$, Michael Gamon$^\ddagger$ \and Henry Kautz$^\dagger$
\\
\\
$^\dagger$Department of Computer Science, University of Rochester \\ 
$^\ddagger$Microsoft Research AI, Microsoft Corporation, Redmond, WA \\
\small{\tt \{nhossain,kautz\}@cs.rochester.edu, \{jckrumm,mgamon\}@microsoft.com}
}

\date{}

\begin{document}
\maketitle


\import{./}{sec0-abstract.tex}
\import{./}{sec1-introduction.tex}

\import{./}{sec2-data.tex}
\import{./}{sec3-task.tex}
\import{./}{sec4-evaluation.tex}
\import{./}{sec5-systems.tex}

\import{./}{sec6-analysis.tex}

\import{./}{sec7-conclusion.tex}

\bibliography{bibliography}
\bibliographystyle{coling}


\end{document}

%% file: sec0-abstract.tex
\begin{abstract}
This paper describes the SemEval-2020 shared task ``Assessing Humor in Edited News Headlines." The task's dataset contains news headlines in which short edits were applied to make them funny, and the funniness of these edited headlines was rated using crowdsourcing.
This task includes two subtasks, the first of which is to estimate the funniness of headlines on a humor scale in the interval 0-3. The second subtask is to predict, for a pair of edited versions of the same original headline, which is the funnier version. To date, this task is the most popular shared computational humor task, attracting 48 teams for the first subtask and 31 teams for the second. 
\end{abstract}

%% file: sec1-introduction.tex
\section{Introduction}
\blfootnote{
    \hspace{-0.65cm}
    This work is licensed under a Creative Commons Attribution 4.0 International License. License details: \url{http://creativecommons.org/licenses/by/4.0/}.
}

\begin{wrapfigure}{R}{0.535\textwidth}
    \centering
    \raisebox{0pt}[\dimexpr\height-2\baselineskip\relax]{
    \begin{tabular}{c}
      \resizebox{0.51\textwidth}{!}{\fbox{
        \includegraphics{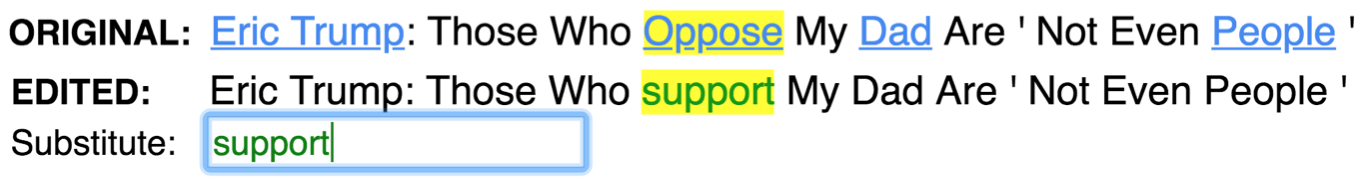}}} \\ 
        \descrcell{\small{(a) The  Headline Editing Interface. }}{\\} \\
      \resizebox{0.51\textwidth}{!}{\fbox{
        \includegraphics{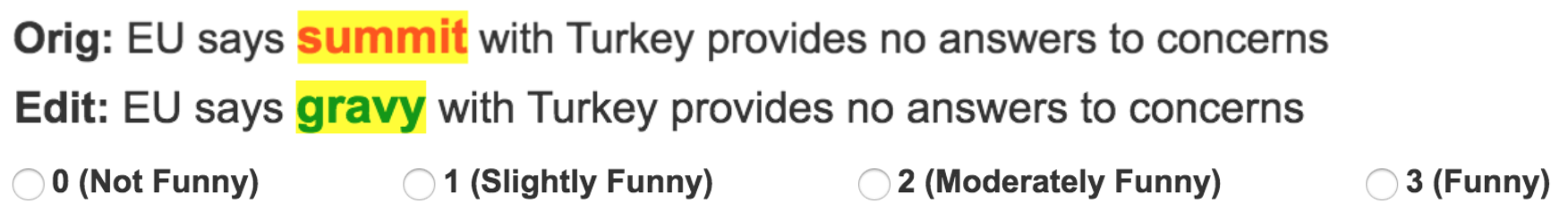}}} \\ \small {(b) The Headline Rating Interface.} \\
    \end{tabular}}
  \caption{The funny headline data annotation interfaces. When editing, only the underlined tokens are replaceable.}
  \label{fig:mturkScreenShots}
\end{wrapfigure}
Humor is an important ingredient of human communication, and every automatic system aiming at emulating human intelligence will eventually have to develop capabilities to recognize and generate humorous content. In the artificial intelligence community, research on humor has been progressing slowly but steadily. As an effort to boost research and spur new ideas in this challenging area, we created a competitive task for automatically assessing humor in edited news headlines.

Like other AI tasks, automatic humor recognition depends on labeled data. Nearly all existing humor datasets are annotated to study the binary task of whether a piece of text is funny~\cite{mihalcea-strapparava-2005-making,kiddon2011s,bertero-fung-2016-long,raz2012automatic,filatova2012irony,zhang2014recognizing,reyes2012humor,barbieri2014automatic}. Such categorical data does not capture the non-binary character of humor, which makes it difficult to develop models that can predict a level of funniness. 

Humor occurs in various intensities, and certain jokes are much funnier than others, including the supposedly funniest joke in the world~\cite{funniestjoke}. A system's ability to assess the degree of humor makes it useful in various applications, such as in humor generation where such a system can be used in a generate-and-test scheme to generate many potentially humorous texts and rank them by funniness, for example, to automatically fill in the blanks in Mad Libs\textsuperscript{\small{\textregistered}} for humorous effects~\cite{hossain2017filling,garimella2020yodalib}. 

For our SemEval 
task, we provided a dataset that contains news headlines with short edits applied to them to make them humorous (see Table~\ref{tab:dataset}). This dataset was annotated as described in Hossain et al.~\shortcite{hossain-etal-2019-president} using Amazon Mechanical Turk, 
where \emph{qualified} human workers edited headlines to make them funny and the quality of humor in these headlines 
was assessed by a separate set of \emph{qualified} human judges on a 0-3 funniness scale (see Figure~\ref{fig:mturkScreenShots}). 
This method of quantifying humor enables the development of systems for automatically estimating the degree of humor in text. Our task is comprised of two Subtasks: 
\begin{itemize}
\itemsep-0.2em 
    \item Subtask 1: Estimate the funniness of an edited headline on a 0-3 humor scale.
    \item Subtask 2: Given two edited versions of the same headline, determine which one is funnier.
\end{itemize}

 
\begin{table}
    \centering
    \resizebox{\linewidth}{!}{
    \begin{tabular}{c|l|l|c|c|c}
        \textbf{ID} & \textbf{Original Headline} (replaced word in \textcolor{red}{\textbf{bold}}) & \textbf{Substitute} & \textbf{Rating} & \textbf{Est.} & \textbf{Err.} \\ \hline
        R1 & CNN 's Jake Tapper to \textbf{\textcolor{red}{interview}} Paul Ryan following retirement announcement & \textcolor{darkGreen}{wrestle} & 2.8 & 1.17 & -1.63  \\ 
        R2 & 4 arrested in Sydney raids to stop terrorist \textcolor{red}{\textbf{attack}}  & \textcolor{darkGreen}{kangaroo} & 2.6
        & 1.06 & -1.54  \\ 
        R3 & Man Sets Off Explosive Device at L.A.-Area Cheesecake Factory, no \textcolor{red}{\textbf{Injuries}}  & \textcolor{darkGreen}{complaints} & 2.4 & 0.80 & -1.60  \\ 
        R4 & 5 dead, 9 injured in \textcolor{red}{\textbf{shooting}} at Fort Lauderdale Airport & \textcolor{darkGreen}{delay} & 1.2 & 0.49 & -0.71 \\ 
        R5 & Congress Struggles to \textcolor{red}{\textbf{Confront}} Sexual Harassment as Stories Pile Up & \textcolor{darkGreen}{increase} & 1.2 & 0.66 & -0.54  \\ 
        R6 & Congress Achieves the Impossible on \textcolor{red}{\textbf{Tax}} Reform & \textcolor{darkGreen}{toilet} & 0.8 & 1.35 & +0.55  \\ 
        R7 & Overdoses now leading \textcolor{red}{\textbf{cause}} of death of Americans under 50 & \textcolor{darkGreen}{sign} & 0.0
        & 0.52 & +0.52 \\ 
        R8 & Noor Salman, widow of Orlando massacre \textcolor{red}{\textbf{shooter}} Omar Mateen, arrested & \textcolor{darkGreen}{columnist} & 0.0 & 0.43 & +0.43 \\ \hline
    \end{tabular}}
    \caption{Edited headlines from our dataset and their funniness rating. We report the mean of the estimated ratings from the top 20 ranked participating systems (Est.) and its difference from the true rating  (Err.).}
    \label{tab:dataset}
\end{table}


Inviting multiple participants to a shared task contrasts with most current work on computational humor, which consists of standalone projects, each exploring a different genre or type of humor. Such projects typically involve gathering new humor data and applying machine learning to solve a particular problem. Repeated attempts at the same problem are rare, hindering incremental progress, which emphasizes the need for unified, shared humor tasks.


Recently, competitive humor tasks including shared data have been posed to the research community. 
One example is \#HashtagWars~\cite{potash2017semeval}, a SemEval task from 2017 that attracted eight distinct teams, where the focus was on ranking the funniness of tweets from a television show. The HAHA competition~\cite{chiruzzo2019overview} had 18 participants who detected and rated humor in Spanish language tweets. There were 10 entries in a SemEval task from 2017 that looked at the automatic detection, location, and interpretation of puns~\cite{miller-etal-2017-semeval}. Finally, a related SemEval 2018 task involved irony detection in tweets~\cite{van2018semeval}. 

Ours is the largest shared humor task to date in terms of participation. More than 300 participants signed up, 86 teams participated in the development phase, and 
48 and 31 teams participated, respectively, in the two subtasks in the evaluation phase. By creating an intense focus on the same humor task from so many points of view, we were able to clearly understand how well these systems work as a function of different dimensions of humor, including which type of humor appears easiest to rate automatically.

%% file: sec2-data.tex
\section{Datasets}
The data\footnote{Task dataset:
\url{https://zenodo.org/record/3969509#.XyWh6fhKh24}}
for this task\footnote{Task competition page: \url{https://competitions.codalab.org/competitions/20970}} is the \textbf{Humicroedit} dataset described in our previous work~\cite{hossain-etal-2019-president}. This dataset contains about 5,000 original headlines, each having three modified, potentially funny versions
for a total of 15,095 edited headlines. The original headlines were collected from Reddit (\texttt{reddit.com}) via the popular subreddits \texttt{r/worldnews} and \texttt{r/politics}, where headlines from professional news sources are posted everyday. These headlines were published between 01/2017 and 05/2018, they are between 4-20 words long, and they are sampled from headlines written by 25 major English news sources.

The data was annotated using workers from Amazon Mechanical Turk, who were screened using a qualification phase to find expert headline editors and judges of humor. 
The editors were instructed to make a headline as funny as possible to a generic wide audience by applying a \textbf{micro-edit}, which is a replacement of a verb/noun/entity in the headline with a single word. Examples are shown in Table~\ref{tab:dataset}. By allowing only small edits, researchers can examine humor at the atomic level where the constrained degrees of freedom are likely to simplify analysis, understanding, and 
eventually generation.


Five judges were asked to rate the funniness of each edited headline using the following humor scale:
\begin{center}
\begin{tabular}{lclclcl}
{\bf 0} - Not funny   &  &  {\bf 1} - Slightly funny & & {\bf 2} - Moderately funny &   & {\bf 3} - Funny
\end{tabular}
\end{center}
\vspace{1pt}
\noindent The funniness of an edited headline is the mean of the ratings from its five judges. For further details and analysis of the dataset, we refer the reader to Hossain et al.~\shortcite{hossain-etal-2019-president}.

For our task, we randomly sampled the Humicroedit dataset into train (64\%), dev (16\%) and test (20\%) sets such that all edited versions of an original headline reside in exactly one of these sets, as opposed to the sampling in Hossain et al.~\shortcite{hossain-etal-2019-president} which allowed overlap of original versions of headlines among its dataset partitions for a slightly different humorous headline classification task.



We also provided additional training data\footnote{FunLines dataset: \url{https://cs.rochester.edu/u/nhossain/funlines.html}} from \textbf{FunLines}\footnote{FunLines game website: \url{https://funlines.co}}~\cite{hossain-etal-2020-funlines}, a competition that we hosted to collect humorous headlines at a very low cost. The data collection approach for Humicroedit and FunLines are mostly similar, but FunLines additionally includes headlines from the news categories sports, entertainment and technology, and its headlines were published between 05/2019 and 01/2020, for a total of 8,248 annotated headlines. 
More than 40\% of the participating teams, including the winning team, made use of the FunLines data. 



%% file: sec3-task.tex
\section{Task Description}
The objective of this shared task is to build systems for rating a humorous effect that is caused by small changes in text. 
To this end, we focus on humor obtained by applying micro-edits to news headlines. 

Editing headlines presents a unique opportunity for humor research since headlines convey  substantial information using only a few words. This creates a rich background against which a micro-edit can lead to a humorous effect. With that data, a computational humor model can focus on the exact localized cause of the humorous effect in a short textual context. 


We split our task into two subtasks. The dataset statistics for these subtasks are shown in Table~\ref{tab:data-stats}. 
\begin{table}
    \centering
    \begin{tabular}{l|l|l|c|c|c|c}
         \textbf{Task} & \textbf{Type} & \textbf{Metric} & \textbf{Train} & \textbf{FunLines (Train)} & \textbf{Dev} & \textbf{Test} \\ \hline
         Subtask 1 & Regression & RMSE & 9,653 & 8,248 & 2,420 & 3,025 \\ 
         Subtask 2 & Classification & Accuracy & 9,382 & 1,959 & 2,356 & 2,961 \\
    \end{tabular}
    \caption{Summary of the subtasks and their datasets.}
    \label{tab:data-stats}
\end{table}

\subsection{Subtask 1: Funniness Regression}
In this task, given the original and the edited versions of a headline, the participant has to estimate the mean funniness of the edited headline on the 0-3 humor scale. 
Systems tackling this task can be useful in a humor generation scenario where generated candidates are ranked according to expected funniness.



\subsection{Subtask 2: Funnier of the Two}
In this task, given the original headline and two of its edited versions, the participating system has to predict which edited version is the funnier of the two. 
Consequently, by looking at gaps between the funniness ratings, we can begin to understand the minimal discernible difference between funny headlines. 



%% file: sec4-evaluation.tex
\section{Evaluation}

\subsection{Metrics}
For Subtask 1, systems are ranked using the root mean squared error (RMSE) between the mean of the five annotators' funniness ratings and the rating estimated by the system for the headlines. Given $N$ test samples, and given the ground truth funniness $y_i$ and the predicted funniness $\hat{y_i}$ for the $i$-th sample:

$$RMSE = \sqrt{\frac{\sum^N_{i=1} (y_i -\hat{y_i})^2}{N}} $$

For Subtask 2, which attempts to find the funnier of the two modified versions of a headline, the evaluation metric is classification accuracy. We also report another auxiliary metric called the reward. Given $N$ test samples with $C$ correct predictions, and given the $i$-th sample, the funniness ratings of its two edited headlines $f_i^{(1)}$ and $f_i^{(2)}$, its ground truth label $y_i$ and its predicted label $\hat{y}_i$:

$$ Accuracy = \frac{C}{N}  \hspace{80pt}  Reward = \frac{1}{N}\sum^N_{i=1}(\mathbbm{1}_{\hat{y_i}=y_i} - \mathbbm{1}_{\hat{y_i}\neq y_i})|f_i^{(1)}-f_i^{(2)}| $$

\noindent In other words, for a larger funniness difference between the two edited headlines in a pair, the reward (or penalty) is higher for a correct classification (or misclassification). We ignore cases where the two edited versions of a headline have the same ground truth funniness. 

\subsection{Benchmarks}
\begin{wraptable}[28]{R}{0.49\textwidth} 
\centering \fontsize{9.0}{12}\selectfont \setlength{\tabcolsep}{0.5em}
\raisebox{0pt}[\dimexpr\height-2\baselineskip\relax]{
\input{results-baselines}}
\caption{Benchmarks on the test set. The best within each model type is \textbf{bolded}, and the overall best is \underline{underlined}.}

\label{tab:baseline}
\end{wraptable}

We provide several benchmarks in Table~\ref{tab:baseline} to compare against participating systems:
\begin{enumerate}
\itemsep-0.1em 
    \item BASELINE: assigns the mean rating (Subtask 1) or the majority label (Subtask 2) from the training set. 
    \item CBOW: the context independent word representations obtained using the pretrained GloVe word vectors with 300d embeddings and a dictionary of 2.2M words. 
    \item BERT: a regressor based on BERT base model embeddings~\cite{devlin-etal-2019-bert}.
    \item RoBERTa: same regressor as above but uses RoBERTa embeddings~\cite{liu2019roberta}. 
\end{enumerate}
\noindent For a thorough discussion of these benchmarks, we refer the reader to the Duluth system \cite{duluth2020semeval}, who performed these ablation experiments. In summary, each benchmark result uses the edited headline, C{\small ONTEXT} implies using the headline's context (with the replaced word substituted with \texttt{[MASK]}), O{\small RIG} implies using the original headline, FT refers to finetuning, F{\small REEZE} implies feature extraction (no finetuning) and F{\small UN}L{\small INES} refers to using the FunLines training data. 

The results for Subtask 2 were obtained by using the model trained for Subtask 1 to assign funniness ratings to both the edited versions of a headline and then choosing the version scoring higher.

\subsection{Results}
The official results for Subtasks 1 and 2 are shown, respectively, in Tables~\ref{tab:res-A} and~\ref{tab:res-B}, including the performance of the benchmarks. 
There were 48 participants for Subtask 1, while Subtask 2 attracted 31 participants. 
For both subtasks, the best performing system was Hitachi, achieving an RMSE of 0.49725 (a 13.5\% improvement over BASELINE) for Subtask 1, 
and an accuracy of 67.43\% (a 17.93 increase in percentage points over BASELINE) for Subtask 2.




%% file: results-baselines.tex
\renewcommand{\arraystretch}{0.88}
\begin{tabular}{@{}l|cH|cHr@{}}
\toprule
{} & \multicolumn{2}{c}{\textbf{Subtask 1}} & \multicolumn{3}{c}{\textbf{Subtask 2}} \\ \textbf{Model} & \textbf{RMSE} & \textbf{Gain} & \textbf{Acc.} & \textbf{Gain} & \textbf{Reward} \\
\midrule
BASELINE & 0.575 & -0.033 & 0.490 & -0.109 & 0.020 \\
\midrule
CBOW &&&&& \\
\quad with \textsc{Context}+\textsc{Freeze} & \textbf{0.542} & \textbf{0.000} &  0.599 & 0.000 & 0.184  \\
\quad +\textsc{Orig} & 0.559 & -0.017 & 0.599 & 0.000 & 0.169 \\
\quad +\textsc{FunLines} & 0.544 & -0.002 & 0.605 & 0.006 & \textbf{0.191} \\
\quad +\textsc{Orig}+\textsc{FunLines} & 0.558 & -0.016 & 0.601 & 0.002 & 0.173 \\
\quad +FT & 0.544 & -0.002 & 0.604 & 0.005 & 0.178 \\
\quad +FT+\textsc{Orig} & 0.561 & -0.019 & 0.592 & -0.007 & 0.165 \\
\quad +FT+\textsc{FunLines} & 0.548 & -0.006 & \textbf{0.606} & \textbf{0.007} & 0.188 \\
\quad +FT+\textsc{Orig}+\textsc{FunLines} & 0.563 & -0.021 & 0.589 & -0.010 & 0.161 \\
\midrule
BERT &&&&& \\
\quad with \textsc{Context}+\textsc{Freeze} & 0.531 & 0.011 & 0.616 & 0.017 & 0.207  \\
\quad +\textsc{Orig} & 0.534 & 0.008 & 0.603 & 0.004 & 0.186 \\
\quad +\textsc{FunLines} & \textbf{0.530} & \textbf{0.012} & 0.615 & 0.016 & 0.207  \\
\quad +\textsc{Orig}+\textsc{FunLines} & 0.541 & 0.001 & 0.615 & 0.016 & 0.204  \\
\quad +FT & 0.536 & 0.006 & \textbf{0.635} & \textbf{0.036} & 0.234  \\
\quad +FT+\textsc{Orig} & 0.536 & 0.006 & 0.628 & 0.029 & 0.231 \\
\quad +FT+\textsc{FunLines} & 0.541 & 0.001 & 0.630 & 0.031 & 0.232 \\
\quad +FT+\textsc{Orig}+\textsc{FunLines} & 0.533 & 0.009 & 0.629 & 0.030 & \textbf{0.236} \\
\midrule
RoBERTa &&&&& \\
\quad with \textsc{Context}+\textsc{Freeze} & 0.528 & 0.014 & 0.635 & 0.036 & 0.246 \\
\quad +\textsc{Orig} & 0.536 & 0.006 & 0.625 & 0.026 & 0.224 \\
\quad +\textsc{FunLines} & 0.528 & 0.014 & 0.640 & 0.041 & 0.252 \\
\quad +\textsc{Orig}+\textsc{FunLines} & 0.533 & 0.009 & 0.618 & 0.019 & 0.207  \\
\quad +FT & 0.534 & 0.008 & 0.649 & 0.050 & \underline{\textbf{0.254}} \\
\quad +FT+\textsc{Orig} & 0.527 & 0.015 & \underline{\textbf{0.650}} & \underline{\textbf{0.051}} & \underline{\textbf{0.254}} \\
\quad +FT+\textsc{FunLines} & 0.526 & 0.016 & 0.638 & 0.039 & 0.233 \\
\quad +FT+\textsc{Orig}+\textsc{FunLines} & \underline{\textbf{0.522}} & \underline{\textbf{0.020}} & 0.626 & 0.027 & 0.216 \\
\bottomrule
\end{tabular}

%% file: sec5-systems.tex
\section{Overview of Participating Systems}
The dominant teams made use of pre-trained language models (\texttt{PLM}), 
namely BERT, RoBERTa, ELMo~\cite{peters-etal-2018-deep}, GPT-2~\cite{radford2019language} and XLNet~\cite{yang2019xlnet}. 
Context-independent word embeddings, such as Word2Vec~\cite{mikolov2013distributed}, FastText~\cite{joulin-etal-2017-bag} and GloVe word vectors~\cite{pennington-etal-2014-glove}, were also useful. 
The winning teams combined the predictions of several hyperparameter-tuned versions of these models using regression in an ensemble learner to arrive at the final prediction. Next, we summarize the top systems and other notable approaches. 

\subsection{Reuse of SubTask 1 System for Subtask 2}
\label{sec:subtask-2-approach}
\begin{wraptable}[44]{r}{0.352\textwidth} 
    \centering
    \raisebox{0pt}[\dimexpr\height-2\baselineskip\relax]{
    \resizebox{0.351\textwidth}{!}{
    \input{results-subtask1}}}
    \caption{Official results and benchmarks for Subtask 1.}
    \label{tab:res-A}
\end{wraptable}
First, we note that for Subtask 2, most systems relied on the model they developed for Subtask 1. This involved using the model to estimate a real number funniness rating for each of the two edited headlines, and selecting the one which achieved the higher estimated rating. As a result, there was a strong correlation between teams' placements in Subtask 1 and Subtask 2, with the top 3 teams in both tasks being the same.

\subsection{The Hitachi System}
The winner of both tasks, Hitachi~\cite{hitachi2020semeval}, formulated the problem as sentence pair regression and exploited an ensemble of the \texttt{PLMs} BERT, GPT-2, RoBERTa, XLNet, Transformer-XL and XLM. Their training data uses the pairs of headlines, with the replacement word marked with special tokens, and they fine-tune 50 instances per \texttt{PLM}, each having a unique hyperparameter setting. After applying 5-fold cross validation, they selected the 20 best performing settings per \texttt{PLM}, for a total of 700 \texttt{PLMs} (7 \texttt{PLMs} $\times$ 20 hyperparameters $\times$ 5 folds). They combined the predictions of these models via Ridge regression in the ensemble to predict final funniness scores. Hitachi uses the additional training data from FunLines. 

\subsection{The Amobee System}
Amobee~\cite{amobee2020semeval} was the 2nd placed team for both Subtasks. Using \texttt{PLM} token embeddings, they trained 30 instances of BERT, RoBERTa and XLNet, combining them for an ensemble of 90 models. 

\subsection{The YNU-HPCC System}
Unlike the top two systems, the 3rd placed YNU-HPCC~\cite{ynu-hpcc2020semeval} employed an ensemble method that uses \emph{only} the edited headlines. They used multiple pre-processing methods (e.g., cased vs uncased, with or without punctuation), and they encoded the edited headlines using FastText, Word2Vec, ELMo and BERT encoders. The final ensemble consists of 11 different encodings (four FastText, two W2V, four Bert, one ELMo). For each of these encodings, a bidirectional GRU was trained using the encoded vectors. In the ensemble, the GRU predictions were concatenated and fed to an XGBoost regressor. 

\subsubsection{MLEngineer}
The MLEngineer~\cite{mlengineer2020semeval} team also used only the edited headlines. They fine-tune and combine four BERT sentence regression models to estimate a rating, and they combine it with the estimated rating from a model that incorporates RoBERTa embeddings and a Na\"ive Bayes regressor to generate the final rating. 


\subsection{The LMML and ECNU Systems}
These systems~\cite{lmml2020semeval,ecnu2020semeval} estimate the funniness of headlines using a neural architecture that focuses on the importance of the replaced and replacement words against the contextual words in the headline.
They use BERT embeddings and compute feature vectors based on the global attention between the contextual words and the replaced (and replacement) word. These two vectors and the vectors of the replaced and replacement are combined, and the resulting vector is passed through a multi-layer perceptron to estimate the headline's funniness.

\subsection{Other Notable Approaches}
\begin{wraptable}{R}{0.435\textwidth} 
\centering
    \raisebox{0pt}[\dimexpr\height-1\baselineskip\relax]{
\resizebox{0.435\textwidth}{!}{
\input{results-subtask2}
}}
    \caption{Official results and benchmarks for Subtask 2.}
    \label{tab:res-B}
\end{wraptable}

ECNU used sentiment and humor lexicons, respectively,  to extract polarities and humor rating features of 
headlines. They also used the average, minimum and maximum humor ratings of replaced/replacement words from the training set as additional features. 

LT3~\cite{lt3-2020semeval} created an entirely feature-engineered baseline which obtained an
RMSE of 0.572. It uses lexical, entity, readability, length, positional, word embedding similarity, perplexity and string similarity features. 

IRLab\_DAIICT trained five BERT classifiers, one for each of the five ratings for a headline, and calculated the mean of the five classifiers' outputs. This mean was further averaged with the output of a BERT regression model which  predicts the overall mean rating.


Buhscitu~\cite{buhscitu2020semeval} used knowledge bases (e.g. WordNet), a language model and hand-crafted features (e.g. phoneme level distances).
Their neural model combines feature, knowledge and word (replaced/replacement) encoders. 

Hasyarasa~\cite{hasyarasa2020semeval} used a word embedding and knowledge graph based approach to build a contextual neighborhood of words to exploit entity interrelationships and to capture contextual absurdity. Features from this and semantic distance based features are finally combined with headline representations from a Bi-LSTM.
 

UTFPR~\cite{utfpr2020semeval} is a minimalist unsupervised approach that uses word co-occurrence features derived from news and EU parliament transcripts to capture unexpectedness.




Some noteworthy pre-processing techniques included non-word symbol removal, word segmentation, manually removing common text extensions in headlines (e.g. ``-- live updates'').
Finally, notable datasets used were the iWeb corpus\footnote{\url{https://www.english-corpora.org/iweb/}} and a news headline corpus\footnote{\url{https://www.kaggle.com/snapcrack/all-the-news}}.

\subsection{General Trends}
Here we discuss the relative merits of the different systems, with respect to the participants' findings. 

Table~\ref{tab:baseline} suggests that contextual information is useful in our humor recognition tasks, since the context independent GloVe embeddings (CBOW) led to weaker performance compared to using the context-sensitive BERT and RoBERTa embeddings. 

According to ablation experiments by Hitachi~\cite{hitachi2020semeval}, the ranking of best performing to least superior individual \texttt{PLM} are as follows: RoBERTa, GPT-2, BERT, XLM, XLNet and Transformer-XL. 

Analysis performed 
by several task participants indicates that
the neural embeddings were unable to recognize humor where a rich set of common sense and/or background knowledge is required, for example, in the case of irony.






Lastly, a few systems had quite low accuracy for Subtask 2. They reported having bugs that caused them to submit a random baseline, which has about a 33\% chance of success (since the possible predictions were ``headline 1 is funnier'', ``headline 2 is funnier'' and ``both headlines have equal funniness'').



%% file: results-subtask1.tex
\begin{tabular}{|c|l|c|}
    \hline 
    \textbf{Rank}  & \textbf{Team}   & \textbf{RMSE} \\ \hline \hline

1  &  Hitachi  &  0.49725 \\ 
2  &  Amobee  &  0.50726 \\ 
3  &  YNU-HPCC  &  0.51737 \\ 
4  &  MLEngineer  &  0.51966 \\ 
5  &  LMML  &  0.52027 \\ 
6  &  ECNU  &  0.52187 \\ \hline
\textbf{bench.} & \textbf{RoBERTa} & \textbf{0.52207} \\ \hline 
7  &  LT3  &  0.52532 \\ 
8  &  WMD  &  0.52603 \\ 
9  &  Ferryman  &  0.52776 \\ 
10  &  zxchen  &  0.52886 \\ \hline
\textbf{bench.} & \textbf{BERT} & \textbf{0.53036} \\ \hline 
11  &  Duluth  &  0.53108 \\ 
12  &  will\_go  &  0.53228 \\ 
13  &  XSYSIGMA  &  0.53308 \\ 
14  &  LRG  &  0.53318 \\ 
15  &  MeisterMorxrc  &  0.53383 \\ 
16  &  JUST\_Farah  &  0.53396 \\ 
17  &  Lunex  &  0.53518 \\ 
18  &  UniTuebingenCL  &  0.53954 \\ \hline
\textbf{bench.} & \textbf{CBOW} & \textbf{0.54242} \\ \hline 
19  &  IRLab\_DAIICT  &  0.54670 \\ 
20  &  O698  &  0.54754 \\ 
21  &  UPB  &  0.54803 \\ 
22  &  Buhscitu  &  0.55115 \\ 
23  &  Fermi  &  0.55226 \\ 
24  &  INGEOTEC  &  0.55391 \\ 
25  &  JokeMeter  &  0.55791 \\ 
26  &  testing  &  0.55838 \\ 
27  &  HumorAAC  &  0.56454 \\ 
28  &  ELMo-NB  &  0.56829 \\ 
29  &  prateekgupta2533  &  0.56983 \\ 
30  &  funny3  &  0.57237 \\ 
31  &  WUY  &  0.57369 \\ 
32  &  XTHL  &  0.57470 \\ \hline
\textbf{bench.} & \textbf{BASELINE} & \textbf{0.57471} \\ \hline 
33  &  HWMT\_Squad  &  0.57471 \\ 
34  &  moonalasad  &  0.57479 \\ 
35  &  dianehu  &  0.57488 \\ 
36  &  Warren  &  0.57527 \\ 
37  &  tangmen  &  0.57768 \\ 
38  &  Lijunyi  &  0.57946 \\ 
39  &  Titowak  &  0.58157 \\ 
40  &  xenia  &  0.58286 \\ 
41  &  Smash  &  0.59202 \\ 
42  &  KdeHumor  &  0.61643 \\ 
43  &  uir  &  0.62401 \\ 
44  &  SO  &  0.65099 \\ 
45  &  heidy  &  0.68338 \\ 
46  &  Hasyarasa  &  0.70333 \\ 
47  &  frietz58  &  0.72252 \\ 
48  &  SSN\_NLP  &  0.84476 \\ \hline
\end{tabular}

%% file: results-subtask2.tex
    \begin{tabular}{|c|l|c|c|}
    \hline 
    \textbf{Rank}  & \textbf{Team}   & \textbf{Accuracy}    & \textbf{Reward} \\ \hline \hline
1  &  Hitachi  & 0.6743 & 0.2988 \\ 
2  &  Amobee  & 0.6606 & 0.2766 \\ 
3  &  YNU-HPCC  & 0.6591 & 0.2783 \\ \hline
\textbf{bench.} & \textbf{RoBERTa} & \textbf{0.6495} & \textbf{0.2541} \\ \hline
4  &  LMML  & 0.6469 & 0.2601 \\ 
5  &  XSYSIGMA  & 0.6446 & 0.2541 \\ 
6  &  ECNU  & 0.6438 & 0.2508 \\ 
7  &  Fermi  & 0.6393 & 0.2438 \\ \hline
\textbf{bench.} & \textbf{BERT} & \textbf{0.6355} & \textbf{0.2345} \\ \hline
8  &  zxchen  & 0.6347 & 0.2399 \\ 
9  &  Duluth  & 0.6320 & 0.2429 \\ 
10  &  WMD  & 0.6294 & 0.2291 \\ 
11  &  Buhscitu  & 0.6271 & 0.2190 \\ 
12  &  MLEngineer  & 0.6229 & 0.2046 \\ 
13  &  LRG  & 0.6218 & 0.2077 \\ 
14  &  UniTuebingenCL  & 0.6183 & 0.2110 \\ 
15  &  O698  & 0.6134 & 0.1954 \\ 
16  &  JUST\_Farah  & 0.6088 & 0.1841 \\ \hline
\textbf{bench.} & \textbf{CBOW} & \textbf{0.6057} & \textbf{0.1878} \\ \hline
17  &  INGEOTEC  & 0.6050 & 0.1779 \\ 
18  &  Ferryman  & 0.6027 & 0.1771 \\ 
19  &  UPB  & 0.6001 & 0.1772 \\ 
20  &  Hasyarasa  & 0.5970 & 0.1673 \\ 
21  &  JokeMeter  & 0.5776 & 0.1487 \\ 
22  &  UTFPR  & 0.5696 & 0.1181 \\ 
23  &  Smash  & 0.5426 & 0.0747 \\ 
24  &  SSN\_NLP  & 0.5377 & 0.0622 \\ 
25  &  WUY  & 0.5320 & 0.1113 \\ 
26  &  uir  & 0.5213 & 0.0567 \\ 
27  &  KdeHumor  & 0.5190 & 0.0272 \\ 
28  &  Titowak  & 0.5038 & -0.0021 \\ \hline
\textbf{bench.}  &  \textbf{BASELINE}	& \textbf{0.4950} &	\textbf{-0.0196}  \\ \hline
29  &  heidy  & 0.4197 & -0.0995 \\ 
30  &  SO  & 0.3291 & -0.2064 \\ 
31  &  HumorAAC  & 0.3204 & -0.2177 \\ \hline

    \end{tabular}

%% file: sec6-analysis.tex
\section{Analysis and Discussion}\label{sec:analysis}

The outputs of 48 participating systems for Subtask 1 and 31 for Subtask 2 present an opportunity to not only study individual solutions and numeric results, but to also take a deeper qualitative look at the output of these systems. Here, we collectively analyze the performance of the top 20 systems per subtask to find aggregate trends that characterize the general approaches and the challenges of assessing humor itself.


\subsection{Subtask 1 (Regression)}
\begin{wrapfigure}{R}{0.5\textwidth}
    \raisebox{0pt}[\dimexpr\height-1\baselineskip\relax]{
  \includegraphics[width=\linewidth]{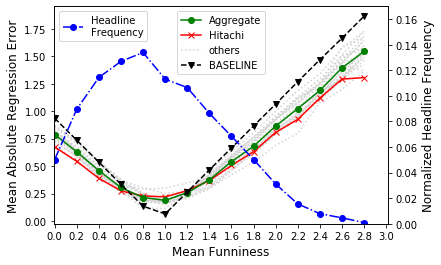}}
  \captionof{figure}{Mean absolute error per funniness bin of width 0.2 for the top 20 systems aggregated, the best system (Hitachi), the 19 other systems and BASELINE for Subtask 1. The blue curve shows the normalized headline frequency for each funniness bin.}
  \label{fig:task1-rmse}
\end{wrapfigure}


To better understand which funniness ranges are particularly hard for systems to assess, we 
study the performance of the systems as a function of ground truth funniness. 
As shown in Figure~\ref{fig:task1-rmse}, we grouped the edited headlines into funniness bins of width 0.2. For each bin, we plotted the mean absolute regression errors for the top 20 systems aggregated (max RMSE = 0.547), the winning Hitachi system (RMSE = 0.497), the 19 other systems and BASELINE (RMSE = 0.575).

In general, all these systems have their minimum error at a funniness score of about $1.0$. While the Hitachi system stands out somewhat in its superior performance at the two extremes of the funniness scale, the other systems follow generally the same pattern, and none appear to be outliers.
Assessing more extreme humor (or lack thereof) appears to be harder since all the systems have larger errors toward the extremes of the funniness scale. This may also be due to the non-uniform distribution of ground truth funniness scores in the dataset (shown as the blue curve), with the extreme values being less frequent.


\subsubsection{Antipodal RMSEs}
\begin{wrapfigure}{R}{0.52\textwidth}
  \raisebox{0pt}[\dimexpr\height-1\baselineskip\relax]{
  \includegraphics[width=\linewidth]{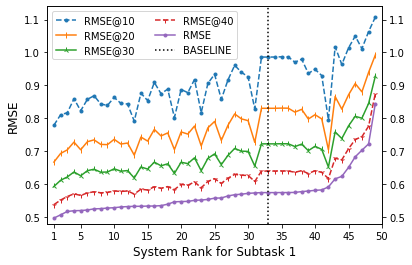}}
  \captionof{figure}{Overall and antipodal RMSE of the ranked participating systems and BASELINE for Subtask 1.}
  \label{fig:task1-rmse-bins}
\end{wrapfigure}
Figure~\ref{fig:task1-rmse-bins} shows the systems' antipodal RMSE, an auxiliary metric for Subtask 1, which we calculated by considering only the $X\%$ most funny headlines and $X\%$ least funny headlines, 
for $X \in \{10,20,30,40\}$ in the RMSE metric. The systems are ranked by their overall RMSE for Subtask 1. It appears that some of the systems further down the ranking are doing much better at estimating the funniness of the extremes in the dataset than their superiors. For example, the large dip shows the system ranked 41 (Hahackathon) is performing better at estimating the funniness of the top 10-40\% most/least funny headlines than several systems ranked before it. This suggests that combining these approaches can yield better results, for example, using some selected systems to rank certain subsets of headlines.

\subsubsection{Systematic Estimation Errors}
We now analyze headlines for which the ratings from the top 20 systems were \emph{all} either underestimates or overestimates. Table~\ref{tab:dataset} shows examples of these headlines, their ground truth funniness rating, the mean of the estimated ratings of the top 20 systems and its difference from the ground truth.

Lack of understanding of world knowledge (Headline R1), cultural references (R2) and sarcasm (R3, R4 and R5) are clearly hurting these systems. The models are having difficulty recognizing the effects of negative sentiments on humor (R7 and R8) and the complex boundaries between negative sentiment and sarcastic humor (R4 and R8 both discuss death but R4 does it in a funny way). 
A better understanding of common sense could have helped resolve these subtleties. R3 also has the humorous effect brought about by a tension relief, which is a complex phenomenon to model. 
Finally, the systems are not expected to infer that bathroom humor (R6) was purposely annotated as ``not funny'' in the data~\cite{hossain-etal-2019-president}.

\subsection{Subtask 2 (Classification)}

\begin{figure}[b]
    \centering
    \resizebox{\linewidth}{!}{
    \begin{tabular}{cc}
        \includegraphics{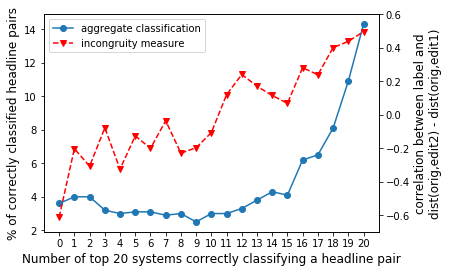} & \includegraphics{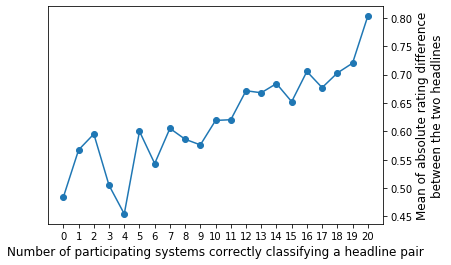} 
    \end{tabular}}
    \begin{tabular}{cc}
        (a) Classification vs. incongruity. & \hspace{60pt} (b) Funniness gaps vs. classification.
    \end{tabular}
    \caption{Aggregate top 20 system classification performance for Subtask 2.}
    \label{fig:task2-analysis}
\end{figure}

Here we examine the top 20 aggregate system performances on Subtask 2.
These 20 systems have at least 59.7\% classification accuracy, much higher than the 49.5\% accuracy of BASELINE.

First, we analyze the difficulty of the classification by calculating the percentage of headline pairs correctly classified by exactly $N$ systems, for $0 \leq N \leq 20$, as shown in the blue curve in Figure~\ref{fig:task2-analysis}(a). 
As an example, there is a subset of about 3\% of the headline pairs that were correctly classified by 10 of the top 20 systems. 
The curve rises rapidly to the right, indicating that a large fraction of the pairs can be correctly classified by 16 or more systems. 

\subsubsection{Incongruity at Play} \label{sec:incongruity}
We investigate to what extent the participating systems model incongruity as a cause of humor, as postulated in the incongruity theory of humor~\cite{sep-humor}. 
This theory claims that jokes set up an expectation that they violate later, triggering surprise and thereby generating humor.
We test this hypothesis by examining the 
cosine distances between the GloVe vectors of the original word and each replacement word. We assume that the larger this distance is, the higher is the expected incongruity.

The dashed curve in Figure~\ref{fig:task2-analysis}(a) shows the \textbf{incongruity measure} obtained using GloVe word distances:
\begin{center}
\vspace{1mm}
\begin{tabular}{|lll|}
\hline 
incongruity difference &=& \texttt{distance(orig, edit$_2$) - distance(orig, edit$_1$)} \\
incongruity measure &=& \texttt{correlation}(incongruity difference, ground truth label $\in \{1,2\}$) \\ \hline
\end{tabular}
\end{center}
\vspace{2mm}
This rising curve implies that the funnier headline in a pair is recognized by more systems if its replacement word is more distant from the original word compared to the distance between the original word and the less funny headline's replacement word.
This indicates that these systems are possibly detecting which headline in the pair is more incongruous 
compared to the original headline.
Moreover, for the headline pairs which were incorrectly classified  by all systems, the incongruity measure is around -0.6, implying that in these headline pairs, the less incongruous (\emph{i.e.,} more coherent) version is the funnier of the two. 
This further indicates that these systems are mostly recognizing incongruity and they tend to fail where incongruity is not the cause of humor.

\subsubsection{Funniness Gaps}
Next, we inspect whether the funniness difference between the two headlines in a pair affects classification accuracy. 
We calculate the mean absolute funniness difference between the headline pairs within each of the $N$ bins of systems that correctly classified them, as shown in Figure~\ref{fig:task2-analysis}(b). For example, the funniness difference between the two headlines in the pairs, which were correctly classified by all 20 systems, was around 0.8 on average. 
The rising trend in the curve suggests that, in general, more systems are able to correctly classify headline pairs having larger differences in humor. This helps confirm the annotation quality in the dataset, showing that humans and machines both agree on the intensity of humor in the dataset, and both can distinguish between slight humor and extreme humor. 
Recall also from Section~\ref{sec:subtask-2-approach} that most of the systems for Subtask 2 were simply applying the systems from Subtask 1 to find the funnier of the two headlines by comparing their funniness scores.
Pairs with widely different funniness would less likely have overlapping uncertainty, leading to more accurate pairwise rankings. 

\subsubsection{Extreme Examples}
\begin{table}
    \centering
    \resizebox{\linewidth}{!}{
    \begin{tabular}{|c|l|l|c|c|c|}
        \hline
         {\bf ID} & {\bf Original Headline} (replaced word in \textbf{\textcolor{red}{bold}}) & {\bf Substitute} & {\bf Rating} & {\bf Dist.}\\ \hline

        \textcolor{darkGreen}{C1} & Secret \textbf{\textcolor{red}{Service}} likely wouldn't have intervened in Trump Jr.-Russia meeting & \textcolor{darkGreen}{police} & 0.0 & 0.72\\ 
        \textcolor{darkGreen}{\ding{52}} & Secret \textbf{\textcolor{red}{Service}} likely wouldn't have intervened in Trump Jr.-Russia meeting & \textcolor{darkGreen}{Santa} & 2.6 & 0.85 \\ \hline
        
        \textcolor{red}{C2} & Amazon, Facebook and Google could save \textbf{\textcolor{red}{billions}} thanks to the GOP tax bill & \textcolor{darkGreen}{puppies} & 1.0 & 0.89 \\ 
        \textcolor{red}{\ding{56}} & Amazon, Facebook and Google could save \textbf{\textcolor{red}{billions}} thanks to the GOP tax bill & \textcolor{darkGreen}{pennies} & 2.2 & 0.54\\ \hline
        
        \textcolor{darkGreen}{C3} & LA Times editorial board condemns Donald Trump \textbf{\textcolor{red}{presidency}} as 'trainwreck' & \textcolor{darkGreen}{diet} & 1.2 & 0.96 \\
        \textcolor{darkGreen}{\ding{52}} & LA Times editorial board \textbf{\textcolor{red}{condemns}} Donald Trump presidency as 'trainwreck' & \textcolor{darkGreen}{celebrates} & 1.0 & 0.69 \\ \hline

        \textcolor{red}{C4} & US officials drop \textbf{\textcolor{red}{mining}} cleanup rule after industry objects & \textcolor{darkGreen}{floor} & 1.4 & 0.86 \\
        \textcolor{red}{\ding{56}} & US officials drop \textbf{\textcolor{red}{mining}} cleanup rule after industry objects & \textcolor{darkGreen}{Bedroom} & 1.2 & 1.01\\ \hline
    \end{tabular}}
    \caption{Examples from Subtask 2 where the top 20 systems collectively either failed (\textcolor{red}{\ding{56}}) or succeeded (\textcolor{darkGreen}{\ding{52}}) in recognizing the funnier headline. On the overall dataset, these were the extreme headline pairs, having either the largest or the smallest differences in funniness between their headlines. We also report the GloVe word vector distances, mapped to the range 0-2, between the replaced and replacement words.}
    \label{tab:task2-qualitative}
\end{table}
We discuss the collective top 20 system performance on edge case examples, with references to Table~\ref{tab:task2-qualitative}:


\begin{itemize}[leftmargin=*]
\itemsep-0.1em 
    \item \textbf{C1:} Among all the test examples which were \emph{correctly} classified by the 20 systems, E1 has the largest funniness difference between its pair of headlines.  ``Secret service'' and ``secret police'' are quite natural in text and substituting one with the other barely changes the headline's meaning. However, using ``secret santa'' clearly raises the surprise. All classifiers were able to assess this relatively easy example. 
    \item \textbf{C2:} This is the example with the largest funniness difference which all 20 systems \emph{incorrectly} classified. 
    This could be because ``puppies'' is semantically more distant from ``billions'' than ``pennies'' (according to GloVe). Although both headline substitutions are funny and incongruous, the antonym effect of the ``pennies'' version triggers a further sarcastic humor, since ``pennies'' is numerically much less than the original word ``billions'', but still in the category of money.  Lacking world knowledge of this numerical difference, the systems award the more incongruous ``puppies'' the higher ranking. As mentioned in~\ref{sec:incongruity}, these systems are especially sensitive to general incongruity as a source of humor and they are likely less aware of other causes of humor, such as meaning reversal.
    \item \textbf{C3:} This example has the smallest funniness difference of the sentences that were correctly classified by all 20 systems. Its less funny headline is sarcastic and most likely all classifiers were unable to recognize sarcasm and thus correctly chose the other headline as the funnier. 
    If this is true, then ignorance about sarcasm was a lucky benefit in this case. 
    \item \textbf{C4:} This was one of the examples with the smallest funniness differences which was misclassified by all systems. Both its headlines are quite funny and they are similar as they both discuss cleaning spaces. However, all systems found bedroom cleaning as a funnier reference than floor cleaning, likely because floor cleaning occurs much more frequently in our day-to-day conversations, making bedroom cleaning a more incongruous substitution to the classifiers, as indicated by the semantic distances in Table~\ref{tab:task2-qualitative}.
\end{itemize}

\subsection{Quirks of the Dataset}

It is challenging to effectively construct a dataset that depends on human creativity, such as humor. Not only generating high quality humor requires more effort from humans making the process expensive, but also reliably assessing the level of humor is challenging as humor understanding is subjective. 

Although we carefully annotated our dataset, we have observed some quirks. Some of our headlines showed lack of sufficient agreement between judges. For example, in the headline \textbf{C2} in Table 6, the standard deviation in judges' ratings for the ``puppies'' version ($\sigma = 0.9$) was much higher than that in the ``pennies'' version ($\sigma = 0.4$), implying that using more judges for the ``puppies'' version could have given it a more reliable funniness rating. 
However, ensuring such quality control would make the data collection process more expensive. 

Additionally, some participating teams reported the frequent mention of President Trump in the dataset, and that there were a non-trivial number of headlines that mentioned both ``Trump'' and ``hair'', and these headlines had received high humor scores, adding certain biases on the data. 

Although the FunLines training data was useful, it was annotated using a different set of judges. It is reasonable to expect that the rating scales of FunLines and our task dataset are not calibrated, and a proper calibration could have possibly increased the value of the FunLines data. However, we have not seen any participating system trying to address this problem, for example, by using a standardization technique to unify the two funniness scales. 


%% file: sec7-conclusion.tex
\section{Conclusion and Future Perspectives}


We provided 15,095 edited and humor-rated, potentially funny headlines and defined subtasks for (1) rating the funniness of each one and (2) determining the funnier headline from a pair that came from editing the same original headline. 
Both humor subtasks were popular, attracting 48 and 31 teams respectively, showing that shared tasks can unify the relatively smaller humor research community. 


For both subtasks, the highly rated solutions show that pre-trained language models work well for rating humor. For Subtask 2, nearly all the participating teams used their solution from Subtask 1 for ranking the two headlines. For Subtask 2, we found that larger disparities in ground truth funniness made ranking easier and that incongruity in a headline was positively correlated with more accurate ranking of humor.
For Subtask 1, we discovered that, over the range of funniness scores, the top systems were most accurate at rating humor near the middle of the funniness range where we had the most training data. 

For future contests like this, we advocate for more uniformly labeled humor data, though that can be hard and expensive to collect. 
Another direction worth pursuing is humor recognition in a closed setting such as reading comprehension, where both annotators and systems make judgments based only on a limited amount of provided contextual information. This would constrain the problem, setting a well-defined scope, and potentially lead to stronger annotator agreements.

We also believe that focusing on specific labeled forms of humor, such as incongruity, sarcasm, irony, puns, and superiority would be advantageous. This could help to better understand how different modeling strategies can identify different root causes of humor.
We would also want to design Subtask 2 to be more independent of Subtask 1 to encourage fresh approaches for Subtask 2. 
Finally, improving the common sense and world knowledge understanding capabilities of AI systems will be crucial for substantially improving the performance of computational humor systems. We hope that both the current results and the dataset in this task provide a stepping stone towards this goal.





\nocite{smash2020semeval,mlengineer2020semeval,lrg2020semeval,xsysigma2020semeval,jokemeter2020semeval,ssn-nlp2020semeval,ferryman2020semeval,elmo-nb2020semeval,hasyarasa2020semeval,hitachi2020semeval,humoraac2020semeval,so2020semeval,kdehumor2020semeval,UniTuebingenCL2020semeval,lmml2020semeval,ecnu2020semeval,amobee2020semeval,wuy2020semeval,ynu-hpcc2020semeval,funny3-2020semeval,buhscitu2020semeval,lt3-2020semeval,utfpr2020semeval,duluth2020semeval}